\documentclass[lettersize,journal]{IEEEtran}
\usepackage[utf8]{inputenc}
\usepackage[T1]{fontenc}
\usepackage{amsmath,amsfonts,amssymb}
\usepackage{algorithmic}
\usepackage{algorithm}
\usepackage{tabularx}
\usepackage{array}
\usepackage{booktabs}    
\usepackage{multirow}    
\usepackage{makecell}   
\usepackage{threeparttable}
\usepackage[table]{xcolor}
\usepackage{textcomp}
\usepackage{dblfloatfix}
\usepackage{verbatim}
\usepackage{graphicx}
\ifCLASSOPTIONcompsoc
  \usepackage[caption=false,font=normalsize,labelfont=sf,textfont=sf]{subfig}
\else
  \usepackage[caption=false,font=footnotesize]{subfig}
\fi

\usepackage{url}
\usepackage{bbding}
\usepackage{orcidlink}
\usepackage{cite}
\usepackage{hyperref}

\hypersetup{
    colorlinks=true,
    linkcolor=blue,
    citecolor=blue,
    urlcolor=black,
}

\makeatletter
\renewcommand\@cite[2]{%
  \textcolor{blue}{[}#1\if@tempswa , #2\fi\textcolor{blue}{]}%
}
\makeatother

\begin{document}

\title{Static and Dynamic Graph Alignment Network for Temporal Video Grounding}

\author{Zhanjie Hu$^{\orcidlink{0009-0000-5561-8463}}$, Bolin Zhang$^{\orcidlink{0009-0006-1728-9682}}$, Jianhua Wang$^{\orcidlink{0000-0001-7488-971X}}$, Jianbo Zheng$^{\orcidlink{0000-0002-7304-1271}}$, Chenchen Yan, \\ Takahiro Komamizu$^{\orcidlink{0000-0002-3041-4330}}$,  \IEEEmembership{Member, IEEE}, Ichiro Ide$^{\orcidlink{0000-0003-3942-9296}}$, \IEEEmembership{Senior Member, IEEE}, Jiangbo Qian$^{\orcidlink{0000-0003-4245-3246}}$, \IEEEmembership{Member, IEEE}%
\thanks{This work was supported in part by Zhejiang Provincial Natural Science Foundation of China under Grant LZ26D010003 and Grant ZCLQN25F0207; in part by NSF, China, under Grant 62271274; in part by Ningbo S\&T Project under Grant 2024Z004; and in part by the K. C. Wong Magna Fund in Ningbo University. \textit{(Corresponding authors: Bolin Zhang and Jiangbo Qian.)}}
\thanks{Zhanjie Hu, and Bolin Zhang are with the Faculty of Electronic Engineering and Computer Science, Ningbo University, Ningbo, China (e-mail: 2411100278@nbu.edu.cn; zhangbolin@nbu.edu.cn).}
\thanks{Jianhua Wang is with the College of Computer Science, Inner Mongolia University, Hohhot, China (e-mail: jhwang@imu.edu.cn).}
\thanks{Jianbo Zheng is with the College of Computing and Engineering, Hunan Normal University, Changsha, China (e-mail: jianbozheng@hunnu.edu.cn).}
\thanks{Chenchen Yan is with Faculty of Computing, Georg-August-Universität Göttingen, Germany, majoring in Applied data science (e-mail: yancc0574@gmail.com).}
\thanks{Takahiro Komamizu is with the Center for Artificial Intelligence, Mathematical and Data Science, Nagoya University, Nagoya, Aichi, Japan (e-mail: taka-coma@acm.org).}
\thanks{Ichiro Ide is with the Graduate School of Informatics, Nagoya University, Nagoya, Aichi, Japan (e-mail: ide@i.nagoya-u.ac.jp).}
\thanks{Jiangbo Qian is with the Merchants' Guild Economics and Cultural Intelligent Computing Laboratory, Ningbo University, Ningbo 315211, China (e-mail: qianjiangbo@nbu.edu.cn).}
}

\markboth{Journal of \LaTeX\ Class Files,~Vol.~XX, No.~X, August~XXXX}%
{Tian \MakeLowercase{\textit{et al.}}: Static and Dynamic Graph Alignment Network for Temporal Video Grounding}

\maketitle

\begin{abstract}
Temporal Video Grounding (TVG) aims to localize temporal moments in an untrimmed video that semantically correspond to given natural language queries.
Recently, Graph Convolutional Networks (GCN) have been widely adopted in TVG to model temporal relations among video clips and enhance contextual reasoning by constructing clip-level graphs. Despite their effectiveness, existing GCN-based TVG methods encounter three critical bottlenecks: 
1) Most methods construct graph nodes using either static or dynamic features alone, resulting in incomplete visual representation and overlooking complementary semantics,
2) Most methods construct temporal graphs in a query-agnostic manner, leading to inefficient feature interaction within the temporal graph representation, and
3) Most methods often suffer from a single-granularity semantic matching, while direct training on complex temporal localization task may lead to slow convergence and suboptimal precision.
To address these challenges, we propose Static and Dynamic Graph Alignment Network (SDGAN). 
First, SDGAN jointly exploits static and dynamic visual features to construct two complementary temporal graphs and performs Position-wise Nodes Alignment, enabling more expressive and robust visual representation.
Second, SDGAN introduces Query--Clip Contrastive Learning and Adaptive Graph Modeling to explicitly align visual clips with their corresponding textual queries, yielding query-aware visual representations.
Third, SDGAN incorporates multi-granularity temporal proposals within Progressive Easy-to-Hard Training Strategy, effectively bridging coarse-grained semantic localization and fine-grained temporal boundary refinement.
Extensive experiments on three benchmark datasets demonstrate that SDGAN achieves superior performance across complex TVG scenarios. 
Codes and datasets are available at \url{https://github.com/ZhanJieHu/SDGAN}.
\end{abstract}


\section{Introduction} 
\label{sec:introduction}
\IEEEPARstart{W}{ith} the continuous advancement of artificial intelligence technology, particularly the widespread application of deep learning methods, fine-grained understanding and organization of video content has become technically feasible~\cite{TVGsurvey2025}. Related research directions such as video question answering~\cite{ VideoQuestionAnswering3}, dense video captioning ~\cite{ DenseVideoCaptioning3}, and temporal action localization~\cite{TemporalActionLocalization3}, have been extensively explored in the academic community. 
As a fundamental task in fine-grained video understanding, Temporal Video Grounding (TVG) is first introduced by Temporal Activity Localization via Language model (TALL)~\cite{TALL} and Moment Context Network (MCN)~\cite{MCN}, where the objective is to retrieve target temporal moments from a long, untrimmed video according to given textual sentences. 
Compared with other video understanding tasks, TVG is particularly challenging due to complex temporal structures, long-range dependencies, and the need for fine-grained alignment between visual events and linguistic semantics~\cite{TVGsurvey2025}.

To effectively model temporal dependencies and contextual relationships among video clips, Graph Convolutional Networks (GCN)~\cite{GCN} have recently been widely adopted in TVG.
GCN-based methods learn node representation by iteratively aggregating information from neighboring nodes defined by a graph structure, enabling effective modeling of relational and contextual patterns~\cite{GCL0}.
This relational reasoning capability makes GCN especially suitable for handling the complex temporal structures inherent in videos~\cite{MOC-CRGMN}.
Consequently, recent TVG methods leverage GCN in various ways, such as constructing clip-level temporal graphs for context modeling~\cite{MOC-CRGMN, UniSDNet}, transforming queries into hierarchical semantic graphs~\cite{chen2020fine, CMIN}, or building scene graphs by recognizing objects and their relationships within video frames~\cite{MKER, DORi}.

\begin{figure}[!t]
	\centering
	
	\subfloat[Existing query-agnostic single-stream GCN-based methods~\cite{MOC-CRGMN, RaNet}.]{
		\includegraphics[width=\columnwidth]{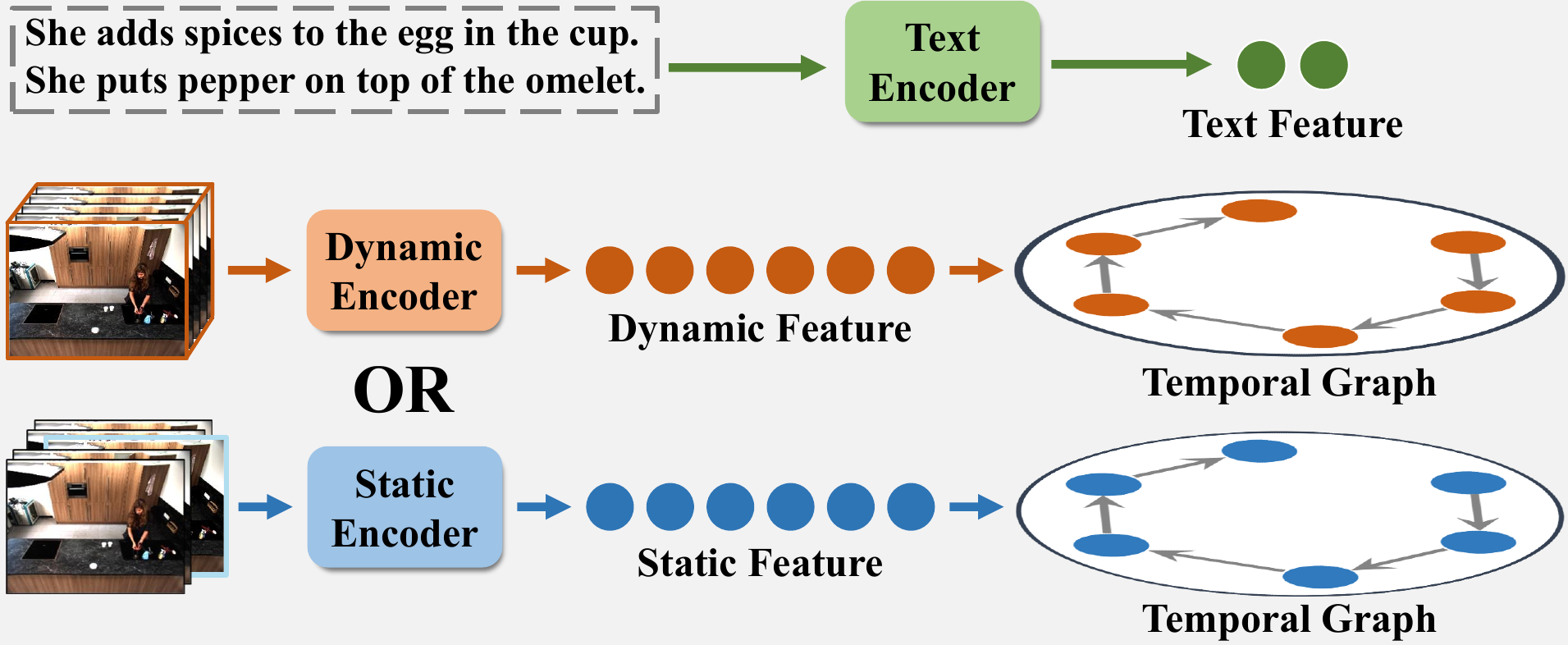}
		\label{fig:1a}
	}

	\subfloat[Static and Dynamic Graph Alignment Network (SDGAN).]{
		\includegraphics[width=\columnwidth]{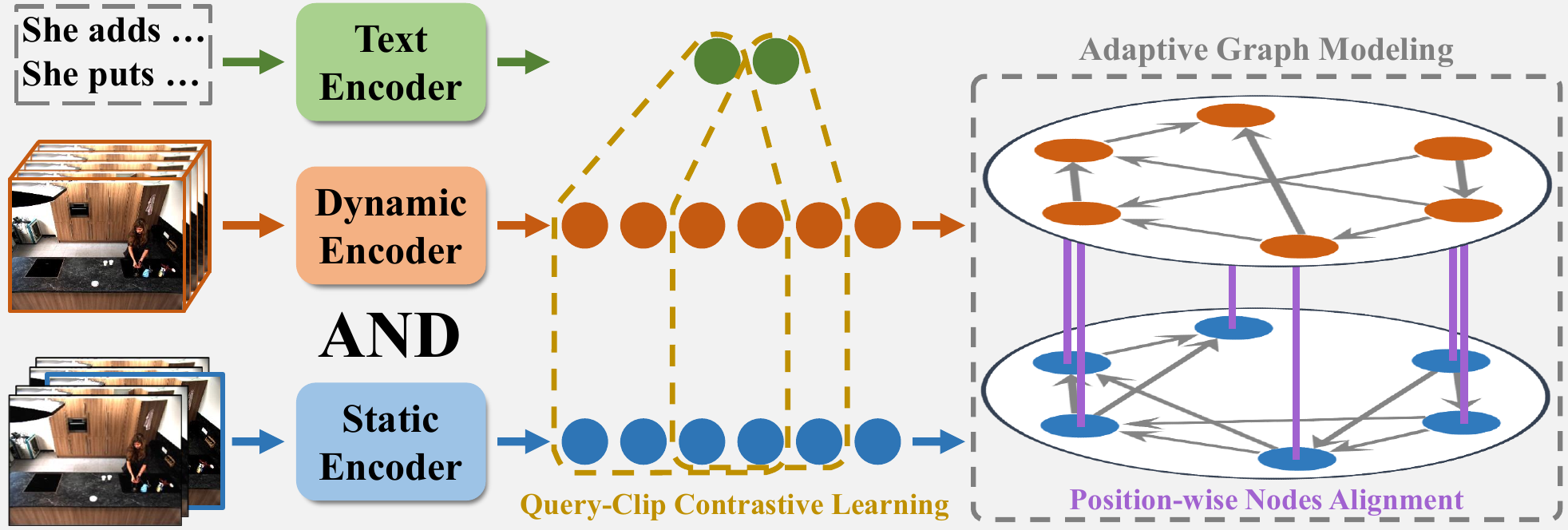}
		\label{fig:1b}
	}
	
	\caption{
	Comparison between existing query-agnostic single-stream graph construction and the proposed SDGAN with static-dynamic graph alignment.
	} 
	\label{fig:paradigm}
\end{figure}

Although GCN-based methods have demonstrated strong potential in improving grounding accuracy, existing methods encounter three critical bottlenecks:
\textbf{\underline{1)}} As illustrated in Fig.~\ref{fig:1a}, existing GCN-based methods predominantly resort to either dynamic~\cite{MOC-CRGMN, UniSDNet} or static~\cite{RaNet, DORi} visual features. 
Specifically, dynamic features are typically derived from 3D ConvNet (C3D)~\cite{C3D} or Two-Stream Inflated 3D ConvNet (I3D)~\cite{I3D}, whereas static features are extracted from sparsely sampled video frames via Contrastive Language--Image Pre-training (CLIP)~\cite{CLIP} or Visual Geometry Group Network (VGGNet)~\cite{VGG}. 
However, relying on a single-modal paradigm inevitably results in incomplete video representation, thereby neglecting the complementary semantic information inherent in videos. 
\textbf{\underline{2)}} Temporal graphs are typically constructed in a query-agnostic manner~\cite{MOC-CRGMN,UniSDNet}, which limits effective feature interaction within temporal graph representations. For instance, Unified Static and Dynamic Network (UniSDNet)~\cite{UniSDNet} constructs temporal graphs based on predefined rules, without explicitly considering the semantic relationship between video content and textual queries.
Therefore, a critical challenge lies in adaptively constructing temporal graphs according to query semantics, rather than relying on predefined rules.
\textbf{\underline{3)}} Existing methods~\cite{MOC-CRGMN,UniSDNet} typically rely on single-granularity temporal proposals, without exploiting the progressive optimization paradigm of curriculum learning, which guides models from easy to hard for improved optimization and generalization~\cite{curriculumlearning}. 
When training is performed solely on fine-grained temporal localization without coarse-grained semantic guidance, the model is prone to unstable optimization and has difficulty establishing reliable temporal boundaries from the outset. 
Moreover, such a single-granularity training strategy may limit robustness to noisy inputs and lead to suboptimal localization performance~\cite{curriculumlearning}.

To this end, we propose \textbf{S}tatic and \textbf{D}ynamic \textbf{G}raph \textbf{A}lignment \textbf{N}etwork (SDGAN) for temporal video grounding. 
\textbf{\underline{First}}, to overcome the limitation of single visual representation, SDGAN employs \textbf{D}ual-\textbf{S}tream \textbf{G}raph \textbf{N}etwork (DSGN) that jointly models static and dynamic features. 
As illustrated in Fig.~\ref{fig:1b}, to mitigate the semantic discrepancy caused by their heterogeneous representation, SDGAN performs \textbf{P}osition-wise \textbf{N}odes \textbf{A}lignment (PNA) between corresponding nodes in dual graphs ({\color[HTML]{a162d0}purple solid lines}).
\textbf{\underline{Second}}, to address the query-agnostic nature of temporal graph construction in existing methods, SDGAN introduces \textbf{Q}uery--\textbf{C}lip \textbf{C}ontrastive \textbf{L}earning (QCCL) and \textbf{A}daptive \textbf{G}raph \textbf{M}odeling (AGM), as illustrated in Fig.~\ref{fig:1b}.
QCCL is employed to align clip features with their corresponding query representations ({\color[HTML]{b95510}brown dashed lines}), explicitly encouraging query-relevant clips to lie closer to the queries in the shared embedding space. 
Such query-aware alignment yields more discriminative node representations for temporal graph modeling.
Building upon these aligned features, SDGAN further performs AGM to enable efficient and targeted feature interaction conditioned on query semantics.
\textbf{\underline{Third}}, to overcome the limitations of single-granularity training strategies, SDGAN adopts a \textbf{P}rogressive \textbf{E}asy-to-\textbf{H}ard \textbf{T}raining strategy (PEHT), inspired by curriculum learning~\cite{curriculumlearning}. 
Specifically, SDGAN alternates the optimization of the coarse-grained and fine-grained branches across training epochs, allowing the model to first localize semantically relevant temporal regions and then progressively refine temporal boundaries. 
By integrating multi-granularity proposal features into this progressive optimization process, SDGAN effectively bridges coarse semantic localization and fine-grained temporal refinement, resulting in improved boundary accuracy and more stable training.

Extensive experiments on three challenging benchmark datasets, namely ActivityNet Caption~\cite{ActivityNetCaptions}, Charades with Sentence Temporal Annotations (Charades-STA)~\cite{TALL}, and Textually Annotated Cooking Scenes (TACoS)~\cite{TACOS}, demonstrate the superior performance of SDGAN.

Our main contributions are summarized as follows:
\begin{itemize}
\item We propose DSGN to jointly model static and dynamic features, and further perform PNA between corresponding nodes across the static and dynamic graphs, thereby enhancing visual representation for TVG.

\item We introduce QCCL and AGM to adaptively construct query-aware temporal graphs, enabling more effective feature interaction between video content and textual queries.

\item We adopt PEHT to progressively bridge coarse-grained semantic localization and fine-grained temporal boundary refinement, leading to improved temporal boundary accuracy and more stable optimization.
\end{itemize}

\section{Related Work}
\label{RW}
\subsection{Temporal Video Grounding}
Current Temporal Video Grounding (TVG) methods can be broadly divided into two mainstream paradigms: Proposal-based and Proposal-free methods~\cite{TVGsurvey2023}. 
From the perspective of architectural design, existing methods can be further categorized into Graph Convolutional Network (GCN)-based methods~\cite{UniSDNet}, Transformer-based methods~\cite{CPL}, and other hybrid frameworks. 
Accordingly, we first review the two mainstream paradigms of TVG, and then focus on representative GCN-based methods.

\subsubsection{\textbf{Proposal-Based and Proposal-Free Paradigms}}
Proposal-based methods generally follow a two-stage generate-and-rank pipeline, in which a set of temporal proposals is first generated and then ranked according to query--proposal matching scores. Early works, such as Temporal Activity Localization via Language model (TALL)~\cite{TALL}~\cite{TALL} and 2D Temporal Adjacent Network (2D-TAN)~\cite{2D-TAN}, rely on sliding windows or 2D temporal maps for proposal generation. More recent methods include Unified Static and Dynamic Network (UniSDNet)~\cite{UniSDNet}, which performs unified feature processing, and Maskable Retentive Network (MRNet)~\cite{MRNet}, which introduces maskable retention to enhance video sequence modeling.
In contrast, proposal-free methods directly regress the start and end timestamps in an end-to-end manner, thereby avoiding the generation of a large number of temporal proposals. However, these methods often rely on aggressive temporal down-sampling, which may result in boundary misalignment~\cite{TVGsurvey2023}. Representative methods include Reaction-Diffusion Multimodal Fusion (RDMF)~\cite{RDMF}, which achieves precise boundary regression through relation-aware multimodal fusion, and Recursive Vision--Language Model (ReVisionLLM)~\cite{ReVisionLLM}, which handles long videos via hierarchical coarse-to-fine recursive refinement.
In this work, we adopt a proposal-based framework and further equip it with a Progressive Easy-to-Hard Training Strategy (PEHT). By progressively optimizing coarse-grained semantic localization and fine-grained temporal boundary refinement, PEHT improves temporal boundary accuracy and promotes more stable training.

\subsubsection{\textbf{GCN-Based Temporal Video Grounding Methods}}
GCN propagates information by aggregating features from topologically neighboring nodes, enabling each node to incorporate local structural context. By stacking multiple graph convolution layers, GCN can further capture higher-order relational dependencies~\cite{GCL0}. Such relational modeling is particularly well suited for TVG, since untrimmed videos often exhibit complex temporal structures, long-range dependencies, and overlapping semantic events~\cite{P-GCN}. By explicitly modeling inter-node temporal relations, GCN provides an effective mechanism for contextual interaction modeling~\cite{MOC-CRGMN}.
Early GCN-based TVG methods~\cite{chen2020fine, CMIN} mainly focus on transforming textual queries into hierarchical semantic graphs while modeling video features in a sequential manner. Subsequent works~\cite{MKER, DORi} further incorporate object detectors to recognize entities in video frames and construct scene graphs for enhanced relational reasoning. More recent methods~\cite{MOC-CRGMN, UniSDNet} introduce clip-level temporal graphs to capture broader contextual dependencies among video clips.
However, these methods~\cite{MOC-CRGMN, UniSDNet} usually construct temporal graphs in a predefined and query-agnostic manner, without explicitly modeling the semantic correspondence between textual queries and video clips. As a result, feature interaction within temporal graph representations is often insufficiently targeted. To address this limitation, we introduce Query--Clip Contrastive Learning (QCCL) and Adaptive Graph Modeling (AGM), which enable efficient and targeted feature interaction conditioned on query semantics.

\subsection{Contrastive Learning}
Contrastive learning~\cite{ContrastiveLearning} has emerged as a dominant representation learning paradigm, largely motivated by the principle of maximum information preservation (Infomax principle)~\cite{GCL82}. Under this principle, the objective is to maximize the Mutual Information (MI)~\cite{MutualInformation} between representations derived from different augmented views of the same instance~\cite{GCL233}. Most early methods~\cite{GCL0} construct positive pairs from augmented views of the same instance and negative pairs from other instances. 
Among them, InfoNCE~\cite{GCL4, GCL114_InfoNCE}, derived from Noise-Contrastive Estimation (NCE)~\cite{GCL36}, is the most widely used objective. It estimates the density ratio and provides a lower bound on MI. It also performs well when sufficient negative samples are available~\cite{GCL231}.
Recent advances have improved contrastive learning through more effective graph augmentations~\cite{GCL233}, refined negative sampling strategies~\cite{GCL177}, and mitigation of false negatives~\cite{GCL41}. Beyond InfoNCE-based objectives, some methods adopt divergence-based objectives, such as Jensen--Shannon Divergence (JSD)~\cite{JSD}, to distinguish joint distributions from marginal distributions, with stabilized variants like Softplus-JSD~\cite{GCL44} improving numerical stability. In addition, some contrastive methods incorporate structural inductive biases~\cite{GCL154, HyperGCL}. For example, Deep Graph Infomax (DGI)~\cite{GCL154} maximizes the mutual information between local patch representations and a global graph summary, thereby emphasizing global--local consistency. 
In this work, SDGAN leverages contrastive learning to align textual and visual information, achieving more precise and robust cross-modal alignment.

\section{Proposed Static and Dynamic Graph Alignment Network (SDGAN)}

\begin{figure*}[!t]
	\centering
	\includegraphics[width=1\textwidth]{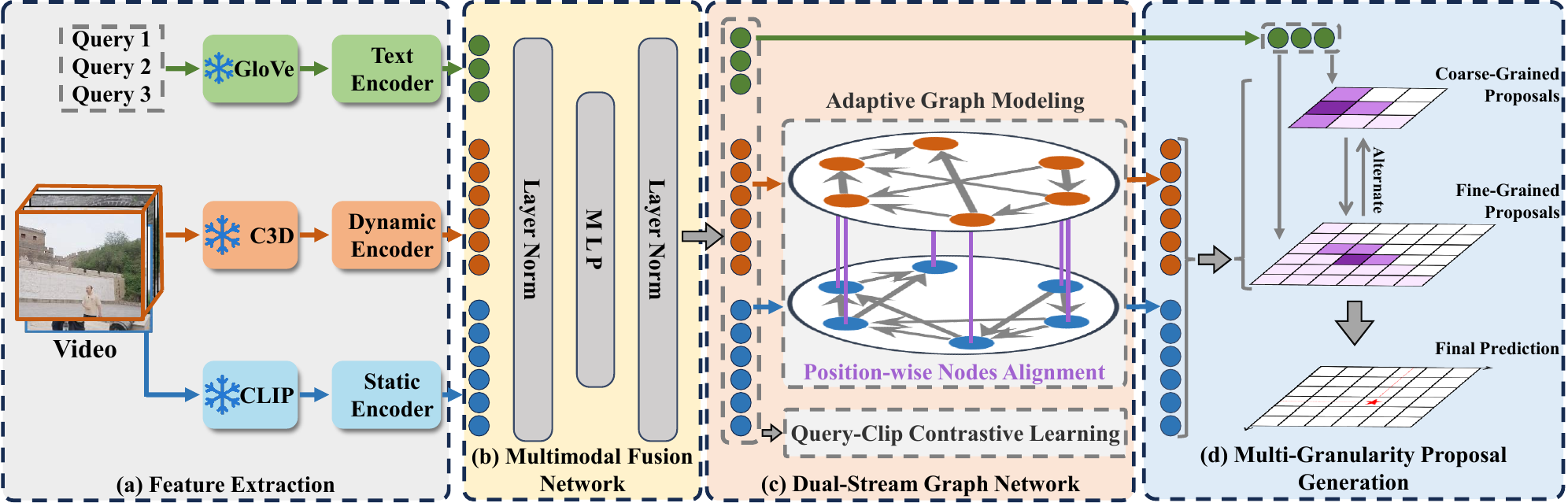}
	\caption{
		\textbf{Overall architecture of the proposed Static and Dynamic Graph Alignment Network (SDGAN).} 
SDGAN consists of four key components: 
(a) Feature Extraction, which extracts dynamic and static visual features from the video together with textual features from the query, 
(b) Multimodal Fusion Network, which aggregates visual and textual features to bridge the semantic gap between the two modalities, 
(c) Dual-Stream Graph Network (DSGN), which constructs static and dynamic temporal graphs in a query-aware manner to enable effective feature interaction, and 
(d) Multi-Granularity Proposal Generation, which produces coarse-grained and fine-grained temporal proposals for moment localization. During training, SDGAN further adopts the Progressive Easy-to-Hard Training strategy (PEHT) to alternately optimize the coarse-grained and fine-grained branches across epochs.
	}
	\label{fig:pic2}
\end{figure*}

\subsection{Task Definition and Framework Overview}
Temporal Video Grounding (TVG) aims to localize the start and end timestamps $(s, e)$ of the target moment in an untrimmed video according to a given natural language query. 
Formally, let the input video be denoted as $V = [f_i]_{i=1}^{F}$, where $F$ is the total number of frames. Given a query $q_i$, the goal is to predict the temporal boundaries $(s_i, e_i)$ of the corresponding moment in $V$. 
During training, each video is associated with a set of annotated query--moment pairs, where the query set is denoted by $Q = \{q_i\}_{i=1}^{N}$ and the corresponding ground-truth moment set is denoted by $M = \{(s_i, e_i)\}_{i=1}^{N}$. Here, $(s_i, e_i)$ represents the start and end timestamps of the target moment corresponding to query $q_i$. During inference, only the video $V$ and query $q_i$ are provided, and the model is required to predict the temporal boundaries of the target moment.

As illustrated in Fig.~\ref{fig:pic2}, SDGAN consists of four main components: Feature Extraction, Multimodal Fusion Network, Dual-Stream Graph Network (DSGN), and Multi-Granularity Proposal Generation. The first three components are designed for multimodal feature processing, while the last component generates temporal proposals at multiple granularities for moment localization.
During training, SDGAN further adopts the Progressive Easy-to-Hard Training strategy (PEHT), which alternates the optimization of the coarse-grained and fine-grained proposal branches across successive epochs.
The following subsections describe the Feature Extraction module (Sec.~\ref{Feature Extraction}), the Multimodal Fusion Network (Sec.~\ref{Multimodal Fusion Network}), the Dual-Stream Graph Network (Sec.~\ref{DSGN}), and the Multi-Granularity Proposal Generation module (Sec.~\ref{Multi-Granularity Proposal Generation}) in detail. The training and inference strategy is presented in Sec.~\ref{Training and Inference}.

\subsection{Feature Extraction}
\label{Feature Extraction}
As shown in Fig.~\ref{fig:pic2}, the original video $V = [f_i]_{i=1}^{F}$ is first divided into clips $[C_k]_{k=1}^{T}$ composed of $T$ frames. 
For dynamic feature extraction, $[C_k]_{k=1}^{T}$ undergoes 3D Convolutional Neural Networks (3D CNN)~\cite{C3D, I3D} to generate a clip-level dynamic feature, which is then fed into the Dynamic Encoder (composed of Conv1D~\cite{CNN} and a linear layer) for mapping, resulting in a dynamic feature $\mathbf{\overline{F}}_{\mathrm{dyn}} \in \mathbb{R}^{T \times D}$. 
For static feature extraction, the input video $[f_i]_{i=1}^{F}$ first undergoes frame downsampling, and the sampled frames are then encoded by an image encoder, i.e., Contrastive Language–Image Pre-training (CLIP)~\cite{CLIP}, followed by the Static Encoder (Video Mamba~\cite{VideoMamba}), to obtain $\mathbf{\overline{F}}_{\mathrm{sta}} \in \mathbb{R}^{T \times D}$.
For the raw query set $Q = \{q_i\}_{i=1}^N$, pretrained word embeddings are first obtained using Global Vectors for word representation (GloVe)~\cite{GloVe}, and are then passed through a Text Encoder composed of linear layers to produce the final query feature $\mathbf{\overline{F}}_{\mathrm{qry}} \in \mathbb{R}^{N \times D}$.
Here, $F$, $T$, and $N$ denote the number of video frames, video clips, and queries, respectively. 
All three types of features are projected to a unified dimension $D$ through their respective encoders, and then serve as inputs to the subsequent Multimodal Fusion Network (Sec.~\ref{Multimodal Fusion Network}).

\subsection{Multimodal Fusion Network} 
\label{Multimodal Fusion Network}

Inspired by Unified Static and Dynamic Network (UniSDNet)~\cite{UniSDNet}, we design a Multimodal Fusion Network to aggregate features from videos and queries.
Dynamic feature $\mathbf{\overline{F}}_{\mathrm{dyn}} \in \mathbb{R}^{T \times D}$, 
static feature $\mathbf{\overline{F}}_{\mathrm{sta}} \in \mathbb{R}^{T \times D}$, and
query feature $\mathbf{\overline{F}}_{\mathrm{qry}} \in \mathbb{R}^{N \times D}$ are first concatenated along the temporal dimension to form a multimodal feature representation as:
\begin{align} 
	\mathbf{\overline{F}}_{\mathrm{VQ}} = [\mathbf{\overline{F}}_{\mathrm{dyn}} \,||\, \mathbf{\overline{F}}_{\mathrm{sta}} \,||\, \mathbf{\overline{F}}_{\mathrm{qry}}] \in \mathbb{R}^{(2T + N) \times D}.
\end{align}

The concatenated feature is subsequently processed by a Multimodal Fusion Network including LayerNorm~\cite{LayerNorm}, Multi Layer Perceptron (MLP), and residual connections~\cite{ResNet} as:
\begin{align} 
	\tilde{\mathbf{F}}_{\mathrm{VQ}} &= \mathbf{\overline{F}}_{\mathrm{VQ}} + \mathrm{LayerNorm}(\mathbf{\overline{F}}_{\mathrm{VQ}}),\\
	\hat{\mathbf{F}}_{\mathrm{VQ}} &= \mathrm{LayerNorm}(\tilde{\mathbf{F}}_{\mathrm{VQ}} + \mathrm{MLP}(\tilde{\mathbf{F}}_{\mathrm{VQ}})).
\end{align}

Finally, the processed feature is split back into dynamic, static, and query components as:
\begin{align} 
	\hat{\mathbf{F}}_{\mathrm{dyn}} &= \hat{\mathbf{F}}_{\mathrm{VQ}}[1 : T, :] \in \mathbb{R}^{T \times D},			\label{eq:Fhat-dyn}\\
	\hat{\mathbf{F}}_{\mathrm{sta}} &= \hat{\mathbf{F}}_{\mathrm{VQ}}[T + 1 : 2T, :] \in \mathbb{R}^{T \times D},		\label{eq:Fhat-sta}\\
	\hat{\mathbf{F}}_{\mathrm{qry}} &= \hat{\mathbf{F}}_{\mathrm{VQ}}[2T + 1 : 2T + N, :] \in \mathbb{R}^{N \times D}.	\label{eq:Fhat-qry}
\end{align}
This design facilitates both intra- and inter-modal semantic interactions among static, dynamic, and query features, enabling the model to capture complementary information and learn richer semantic representations~\cite{UniSDNet}.

\subsection{Dual-Stream Graph Network (DSGN)}
\label{DSGN}
Given the clip-level video feature $\hat{\mathbf{F}}_m \in \mathbb{R}^{T \times D}$ ($m \in \{\mathrm{dyn}, \mathrm{sta}\}$) (Eq.~\ref{eq:Fhat-dyn} and~\ref{eq:Fhat-sta}) produced by the Multimodal Fusion Network (Sec.~\ref{Multimodal Fusion Network}), we aim to explicitly embed query semantics into the video feature. 
To this end,  DSGN is performed in a synchronous manner on both dynamic and static features, where the subscript $m$ denotes either the dynamic or static modality.
The entire pipeline proceeds as follows:
Given $\hat{\mathbf{F}}_{m}$ produced by the Multimodal Fusion Network (Sec.~\ref{Multimodal Fusion Network}), Query--Clip Contrastive Learning (Sec.~\ref{QCCL}) is applied to enhance query-aware discriminability, while Adaptive Graph Modeling (Sec.~\ref{AGM}) adaptively models feature interactions over $\hat{\mathbf{F}}_{m}$ to enable targeted information propagation.

\subsubsection{\textbf{Query--Clip Contrastive Learning (QCCL)}}
\label{QCCL}
\begin{figure}[!t]
	\centering
	\includegraphics[width=\columnwidth]{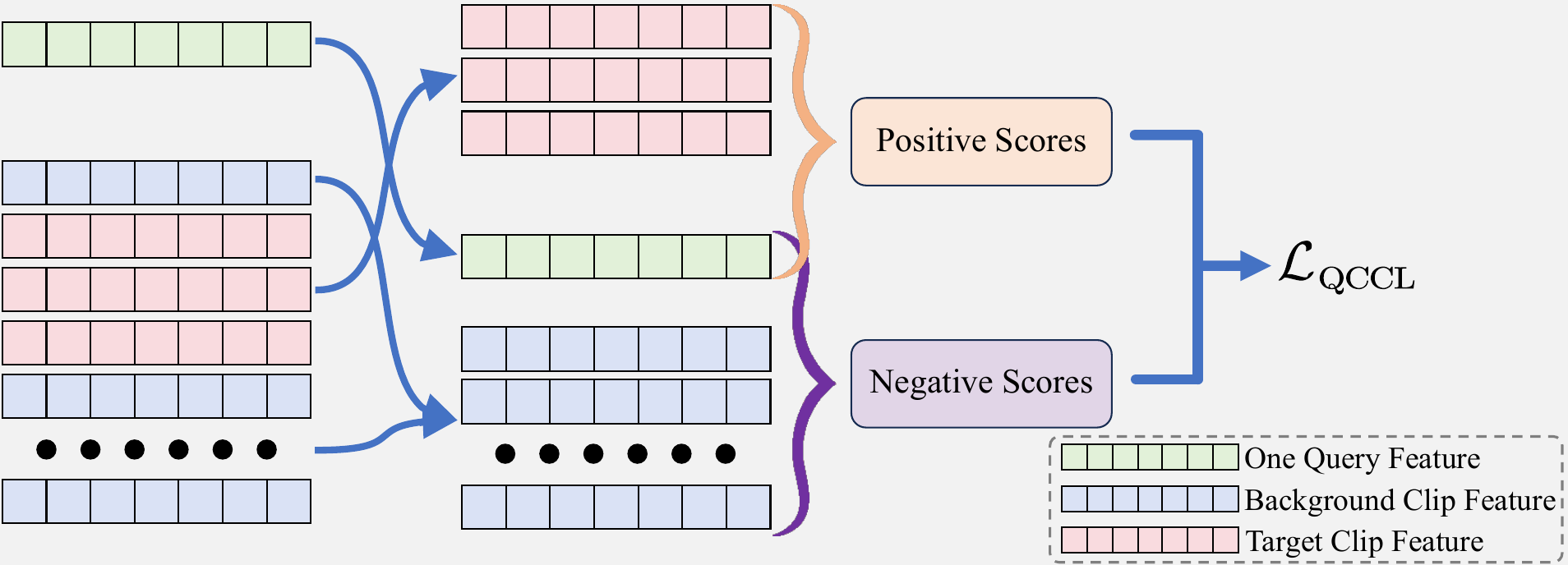}
	\caption{Proposed Query--Clip Contrastive Learning (QCCL) brings relevant query--clip pairs closer together while pushing irrelevant ones apart.}
	\label{fig:pic3}
\end{figure}

Inspired by Zhang et al.'s work~\cite{ReLoCLNet}, QCCL aligns video and query features by bringing relevant query--clip pairs closer together while pushing irrelevant ones apart, thereby achieving discriminative semantic alignment.
The schematic diagram is illustrated in Fig.~\ref{fig:pic3}.
As defined previously, $\hat{\mathbf{F}}_{\mathrm{qry}} = [\mathbf{q}_i]_{i=1}^N$ (Eq.~\ref{eq:Fhat-qry}) denote the encoded features of $N$ queries, 
and $\hat{\mathbf{F}}_{m} = [\mathbf{f}_i]_{i=1}^T$ denote the features of $T$ video clips. 
For each query $\mathbf{q}_i$, the clip features whose temporal spans are fully contained within the target moment are regarded as positive samples, denoted as $\mathbf{\hat{F}}_{i,\mathrm{P}} \subseteq \hat{\mathbf{F}}_m$.
Meanwhile, the remaining clip features are regarded as negative samples, denoted as $\mathbf{\hat{F}}_{i,\mathrm{N}} = \hat{\mathbf{F}}_m \setminus \mathbf{\hat{F}}_{i,\mathrm{P}}$.
The QCCL loss $\mathcal{L}_{\mathrm{QCCL}}$ is formulated as:
\begin{equation} 
	\begin{split}
		\mathcal{L}_{\mathrm{QCCL}} = -\frac{1}{N} \sum_{i=1}^N \Bigg[ 
		&\mathrm{E}_{\mathbf{f} \in \mathbf{\hat{F}}_{i,\mathrm{P}}} \Bigl[-\mathrm{sp}\bigl(-\mathrm{C}(\mathbf{q}_i, \mathbf{f})\bigr) \Bigr] \\
		&- \mathrm{E}_{\mathbf{f} \in \mathbf{\hat{F}}_{i,\mathrm{N}}} \Bigl[ \mathrm{sp}\bigl(\mathrm{C}(\mathbf{q}_i, \mathbf{f})\bigr) \Bigr]\Bigg],
		\label{eq:QCCL} 
	\end{split}
\end{equation}
where $\mathrm{C}(\cdot, \cdot)$ denotes a discriminator that takes a query vector and a video feature vector as inputs, and outputs a scalar score measuring their semantic correlation.
\( \mathrm{sp}(x) = \log(1 + e^{x}) \) is Softplus activation, 
and $\mathrm{E}[\cdot]$ denotes the expectation.
This loss maximizes mutual information~\cite{MutualInformation} between queries and their matching clips while suppressing irrelevant correlations, guiding the model to identify query-corresponding clips.

By enforcing QCCL, the preceding Multimodal Fusion Network (Sec.~\ref{Multimodal Fusion Network}) can generate a query-aware discriminative video feature. This significantly improves the feature quality and provides a strong foundation for selective inter-node interaction in AGM (Sec.~\ref{AGM}).

\subsubsection{\textbf{Adaptive Graph Modeling (AGM)}}
\label{AGM}
Based on query-aware feature $\hat{\mathbf{F}}_{m} \in \mathbb{R}^{T \times D}$ where $m \in \{\mathrm{dyn}, \mathrm{sta}\}$ (Eqs.~\ref{eq:Fhat-dyn} and \ref{eq:Fhat-sta}), AGM constructs static and dynamic temporal graphs by mapping each clip-level feature to a node.
Within the $A$ layers of both graphs, AGM establishes selective and adaptive edges, facilitating a more flexible and robust interaction while suppressing noise.

\textbf{Graph Construction.}
We consider potential edges between any pair of nodes.
To preserve temporal consistency, directed edges are constructed from nodes with smaller temporal indices to those with larger ones.
The graph is then sparsified by retaining the top-$K$ directed edges with the highest cosine similarity among all valid candidate pairs.
This adaptive connectivity enables effectively modeling the semantic dependencies within the video representation.

\textbf{Graph Update.} 
Each node feature $\mathbf{f}_i$ is updated by aggregating information from its neighborhood $\mathcal{N}(i) = \{ \mathbf{f}_j \mid j \in \text{adj}(i) \}$. 
Formally, the update process for $\mathbf{f}_i$ at the $a$-th graph layer is formulated as:
\begin{align}
\mathbf{f}_i^{(a+1)} = \sigma ( \sum_{\mathbf{f}_j\in N(\mathbf{f}_i)} \Phi (i-j) \cdot \mathbf{f}_j^{(a)} ) 
\end{align}
where $\Phi(\cdot)$ refers to a Radial Basis Function~\cite{RBF}, and 
$\sigma(\cdot)$ introduces non-linearity after information propagation.
Moreover, $i-j$ represents the temporal distance between nodes $i$ and node $j$, and is used as the edge weight to incorporate temporal information into the graph.

AGM yields the final representation $\mathbf{F}_{m} = [\mathbf{f}_i^{(A)}]_{i=1}^T$, where $m \in \{\mathrm{dyn}, \mathrm{sta}\}$ capturing semantically meaningful long-range dependencies.

\subsection{Multi-Granularity Proposal Generation}
\label{Multi-Granularity Proposal Generation}
Most existing methods~\cite{2D-TAN, UniSDNet} rely on single-granularity semantic matching, while directly optimizing fine-grained temporal localization often leads to slow convergence and suboptimal precision.
Inspired by curriculum learning principles~\cite{curriculumlearning}, SDGAN incorporates multi-granularity temporal proposals effectively bridging coarse-grained semantic localization and fine-grained temporal boundary refinement.
This section details the Multi-Granularity Proposal Generation module, which produces multi-granularity proposals in a synchronous manner on both dynamic feature ${\mathbf{F}}_{\mathrm{dyn}}$ and static feature ${\mathbf{F}}_{\mathrm{sta}}$ via three steps: 
1) Multi-Granularity Video Feature Aggregation (Sec.~\ref{Multi-Granularity Video Feature Aggregation}) for hierarchical temporal representation, 
2) Multi-Granularity Proposal Feature Construction (Sec.~\ref{Multi-Granularity Proposal Feature Construction}) to build coarse- and fine-grained proposal features, and  
3) Proposal-Query Matching (Sec.~\ref{Proposal-Query Matching}) to score cross-modal relevance and guide training process.
This pipeline effectively supports PEHT (Sec.~\ref{Progressive Easy-to-Hard Training Strategy (PEHT)}).

\subsubsection{\textbf{Multi-Granularity Video Feature Aggregation}}
\label{Multi-Granularity Video Feature Aggregation}
For each visual stream $m \in \{\mathrm{dyn}, \mathrm{sta}\}$, we first construct a coarse-grained video feature by uniformly aggregating consecutive clip-level features.
Given the original fine-grained feature ${\mathbf{F}}_{m} \in \mathbb{R}^{T \times D}$, we apply a non-overlapping temporal window of size $n$ with a stride of $n$, where every $n$ consecutive clip features are aggregated into one single coarse-grained feature.
This operation yields a lower-resolution video feature sequence ${\mathbf{FL}}_{m} \in \mathbb{R}^{t \times D}$, with the constraint $t \times n = T$.

\subsubsection{\textbf{Multi-Granularity Proposal Feature Construction}}
\label{Multi-Granularity Proposal Feature Construction}
We generate the proposal feature in a synchronous manner on both fine-grained clip-level feature ${\mathbf{F}}_{m} \in \mathbb{R}^{T \times D}$ and coarse-grained feature ${\mathbf{FL}}_{m} \in \mathbb{R}^{t \times D}$.
For fine-grained feature $\mathbf{F}_m \in \mathbb{R}^{T \times D}$, we adopt moment sampling~\cite{2D-TAN, UniSDNet} to construct a 2D temporal proposal map. 
Specifically, for each pair of clip features $\mathbf{f}_i$ and $\mathbf{f}_j$ with $i \le j$, we compute the proposal feature $\mathbf{m}_{ij}$ as:
\begin{equation} 
	\mathbf{m}_{ij} = \operatorname{MaxPool}(\mathbf{f}_i, \dots, \mathbf{f}_j) + \mathbf{f}_i + \mathbf{f}_j \in \mathbb{R}^D.
\end{equation}
These features are then arranged into an initial 2D map $\mathbf{M}^\mathrm{2D}_\mathrm{init} \in \mathbb{R}^{T \times T \times D}$, which are further processed by convolution~\cite{CNN} to obtain the final proposal representations $\mathbf{M}^\mathrm{2D} \in \mathbb{R}^{T \times T \times D}$.

Similarly, based on the coarse-grained feature  ${\mathbf{FL}}_{m} \in \mathbb{R}^{t \times D}$, we generate a lower-resolution 2D proposal map $\mathbf{ML}^\mathrm{2D} \in \mathbb{R}^{t \times t \times D}$.

\subsubsection{\textbf{Proposal--Query Matching}}
\label{Proposal-Query Matching}
Taking fine-grained feature as an illustrative example, to quantify the relevance between $ N $ queries and proposals, 
we first project the 2D temporal proposal map $\mathbf{M}^\mathrm{2D} \in \mathbb{R}^{T \times T \times D}$ and the query feature $\hat{\mathbf{F}}_{\mathrm{qry}} \in \mathbb{R}^{N \times D}$ into normalized vectors. 
Then we calculate their cosine similarity, following the paradigm of UniSDNet~\cite{UniSDNet} as:
\begin{equation}
	\begin{split}
		\mathbf{S}
		&= \mathrm{Cos}\Bigl( \mathrm{Norm}\bigl(\mathbf{M}^\mathrm{2D}\bigr),
		\mathrm{Norm}\bigl(\hat{\mathbf{F}}_{\mathrm{qry}}\bigr) \Bigr) \\
		&= \{s_{ijk}\} \in \mathbb{R}^{N \times T \times T},
	\end{split}
	\label{eq:modality_alignment}
\end{equation}
where $\mathrm{Norm}(\cdot)$ denotes Normalization, and
$s_{ijk}$ denotes the relevance score between the $i$-th query ($i \in [1,N]$) and the proposal that starts at clip $j$ and ends at clip $k$ ($j,k \in [1,T]$).
For each query $\mathbf{q}_i$, the best-matching moment is the proposal with the highest score.
The same procedure is applied to the coarse-grained 2D proposal map $\mathbf{ML}^\mathrm{2D} \in \mathbb{R}^{t \times t \times D}$ to obtain the corresponding coarse-grained score map $\mathbf{SL} \in \mathbb{R}^{N \times t \times t}$.

\begin{table*}[t]
	\centering
	\caption{Hyperparameter configurations and training settings of the proposed Static and Dynamic Graph Alignment Network (SDGAN). 
DSGN refers to the Dual-Stream Graph Network (Sec.~\ref{DSGN}). 
Hidden denotes the hidden-layer dimension. 
Coarse and Fine denote the numbers of coarse- and fine-grained clips, respectively. 
Max indicates that coarse clips are generated from fine-grained clips via max pooling.}
	\label{tab:hyperparam_training}
	\resizebox{\textwidth}{!}{
		\begin{tabular}{@{}l c c crc cc rrr@{}}
			\toprule
			\multirow{2}{*}{\textbf{Dataset}} & \multirow{2}{*}{\textbf{Dynamic Feature}} & \multirow{2}{*}{\textbf{Hidden}} & \multicolumn{3}{c}{\textbf{Num of Clips}} & \multicolumn{2}{c}{\textbf{DSGN}} & \multicolumn{3}{c}{\textbf{Training Strategy}} \\
			
			\cmidrule(lr){4-6} \cmidrule(lr){7-8}  \cmidrule(lr){9-11}
			
			& & & \textbf{Coarse} & \textbf{Fine} & \textbf{Method} & \textbf{Layers} & \textbf{Edges} & \textbf{Learning Rate} & \textbf{Batch Size} & \textbf{Epochs} \\
			\midrule
			ActivityNet Caption~\cite{ActivityNetCaptions} & C3D~\cite{C3D} & 256 & 32 & 64  & Max & 2 & 1,104 & $8 \times 10^{-4}$ & 32 & 32 \\
			Charades-STA~\cite{TALL} 	& I3D~\cite{I3D} & 256 & 64 & 128  & Max & 2 & 1,104 & $1 \times 10^{-3}$ & 16 & 20 \\
			TACoS~\cite{TACOS} 		& C3D~\cite{C3D} & 256 & 64 & 128 & Max & 2 & 2,548 & $15 \times 10^{-4}$ & 4 & 140 \\
			\bottomrule
		\end{tabular}
	}
\end{table*}

\subsection{Training and Inference}
\label{Training and Inference}
We adopt multiple loss functions for model optimization. 
Meanwhile, PEHT is leveraged in the training phase.

\subsubsection{\textbf{Loss Functions}}
\label{loss functions}
The proposed method jointly optimizes four different yet complementary losses.

\textbf{First}, the proposed Query--Clip Contrastive Learning loss $\mathcal{L}_{\mathrm{QCCL}}$ (Eq.~\ref{eq:QCCL}) encourages the Multimodal Fusion Network to generate query-aware and discriminative video features, thereby providing a strong foundation for selective inter-node interaction in AGM.

\textbf{Second}, after Multi-Granularity Video Feature Aggregation (Sec.~\ref{Multi-Granularity Video Feature Aggregation}), we aim to mitigate the disruptions arising from intra-modal noise and temporal misalignment.
To this end, we introduce \textbf{Position-wise Nodes Alignment (PNA)} loss $\mathcal{L}_{\mathrm{PNA}}$ on static and dynamic nodes through Information Noise-Contrastive Estimation (InfoNCE) loss~\cite{GCL114_InfoNCE} as:
{
\begin{align}
	\mathcal{L}_{\mathrm{PNA}} = -\frac{1}{T} \sum_{t=1}^{T} 
	\log\frac
	{\exp\left(\mathrm{Cos}\left(\mathbf{f}_{\mathrm{dyn},t},
		\mathbf{f}_{\mathrm{sta},t}\right)/\tau\right)}
	{\sum_{t'=1}^{T} \exp\left(\mathrm{Cos}\left(\mathbf{f}_{\mathrm{dyn},t},
		\mathbf{f}_{\mathrm{sta},t'}\right)/\tau\right)},
	\label{eq:PNA}
\end{align}
}where $\mathbf{f}_{\mathrm{dyn},t}$ and $\mathbf{f}_{\mathrm{sta},t}$ are the $t$-th position node of $\mathbf{F}_{\mathrm{dyn}}$ and $\mathbf{F}_{\mathrm{sta}}$, respectively, and $\tau$ is the temperature coefficient. This loss aligns the dynamic and static nodes that share the same position. 
Through the aforementioned pipeline, DSGN explicitly injects query semantics into video representations, thereby enabling targeted information propagation and more discriminative feature learning. 
As detailed in Sec.~\ref{Multi-Granularity Video Feature Aggregation}, video features are constructed at two granularity levels. Accordingly, the PNA loss is defined with two variants, namely the coarse-grained loss $\mathcal{L}_{\mathrm{PNA,coarse}}$ and the fine-grained loss $\mathcal{L}_{\mathrm{PNA,fine}}$.

\textbf{Third}, the score maps $\mathbf{S}$ and $\mathbf{SL}$ are optimized using an Intersection over Union (IoU) regression objective. 
Following UniSDNet~\cite{UniSDNet}, we compute the ground-truth IoU values $\mathrm{IoU}^{\mathrm{GT}} = \{\mathrm{IoU}_i\}_{i=1}^{M} \in \mathbb{R}^{M}$, where $M$ denotes the total number of score under consideration, and each $\mathrm{IoU}_i \in [0,1]$ is the IoU between the proposal and target ground-truth moments $(s_i, e_i)$. 
The IoU regression loss $\mathcal{L}_{\mathrm{IoU}, m}$ is then formulated as:
{
\begin{equation}
	\mathcal{L}_{\mathrm{IoU}, m} = \frac{1}{M} \sum_{i=1}^{M} \left[ \mathrm{IoU}_i \log y_i + (1 - \mathrm{IoU}_i) \log (1 - y_i) \right], \label{eq:iou_loss}
\end{equation}
}where $m \in \{\mathrm{dyn}, \mathrm{sta}\}$ and $y_i$ represents the model's predicted IoU score for the proposal, extracted from the score map $\mathbf{S}$ (or $\mathbf{SL}$) as described in Eq.~\ref{eq:modality_alignment}.

\textbf{Fourth}, we design the query-proposal contrastive learning loss $\mathcal{L}_{\mathrm{Contra}}$ on both fine-grained and coarse-grained 2D proposal feature.
Based on Noise-Contrastive Estimation (NCE)~\cite{GCL36}, $\mathcal{L}_{\mathrm{Contra}}$ models two conditional matching distributions: $p(q_m|m)$ (probability of a query $q_m$ matching a proposal $m$ given $m$) and $p(m_q|q)$ (probability of a proposal $m_q$ matching a query $q$ given $q$). It is defined as:
{
\begin{equation}
	\mathcal{L}_{\mathrm{Contra}, m'} = -\Bigg( \sum_{q \in Q^B} \log p(m_q|q) + \sum_{m \in M^B} \log p(q_m|m) \Bigg), \label{eq:contra_loss}
\end{equation}
}where $m' \in \{\mathrm{dyn}, \mathrm{sta}\}$, $Q^B$ and $M^B$ denote the query set and proposal set associated with a video, respectively.
This loss guides effective representation learning for both query and video features.

Accordingly, the overall loss is defined as:
\begin{equation}
\begin{aligned}
\mathcal{L} =\;& \lambda_{\mathrm{Q}} \mathcal{L}_{\mathrm{QCCL}}
+ \lambda_{\mathrm{C}} \mathcal{L}_{\mathrm{PNA,\,coarse}}
+ \lambda_{\mathrm{F}} \mathcal{L}_{\mathrm{PNA,\,fine}} \\
&+ \lambda_{\mathrm{D}} \left(
\mathcal{L}_{\mathrm{IoU,\,dyn}} +
\mathcal{L}_{\mathrm{Contra,\,dyn}}
\right) \\
&+ \lambda_{\mathrm{S}} \left(
\mathcal{L}_{\mathrm{IoU,\,sta}} +
\mathcal{L}_{\mathrm{Contra,\,sta}}
\right)
\end{aligned}
\label{eq:final_loss}
\end{equation}
where $\lambda_\mathrm{Q}$, $\lambda_\mathrm{C}$, $\lambda_\mathrm{F}$, $\lambda_\mathrm{D}$, and $\lambda_\mathrm{S}$ are balancing coefficients for the corresponding loss terms. Based on the ablation studies in Sec.~\ref{Ablation Studies}, the best performance is achieved with $\lambda_\mathrm{Q}=1.0$, $\lambda_\mathrm{C}=1.0$, $\lambda_\mathrm{F}=0.0$, $\lambda_\mathrm{D}=0.6$, and $\lambda_\mathrm{S}=0.4$.

\subsubsection{\textbf{Progressive Easy-to-Hard Training Strategy (PEHT)}}
\label{Progressive Easy-to-Hard Training Strategy (PEHT)}
Inspired by curriculum learning principles~\cite{curriculumlearning}, we incorporate PEHT, which effectively bridges coarse-grained semantic localization and fine-grained temporal boundary refinement.
During training, we first use the coarse-grained proposal map $\mathbf{ML}^{\mathrm{2D}}$ (Sec.~\ref{Multi-Granularity Proposal Feature Construction}) and its score map $\mathbf{SL}$ (Sec.~\ref{Proposal-Query Matching}) for initial training, enabling the model to learn rough query--video semantic alignment. 
Then, we introduce the fine-grained $\mathbf{M}^{\mathrm{2D}}$ (Sec.~\ref{Multi-Granularity Proposal Feature Construction}) and $\mathbf{S}$ (Eq.~\ref{eq:modality_alignment}) for optimization, refining temporal boundary accuracy.
Furthermore, we alternate between two training stages across successive epochs.
The proposed PEHT guides the model through a smoother optimization path.
Coarse-grained supervision first builds rough temporal representation to broadly localize candidate regions, while fine-grained stages subsequently sharpen precise boundary localization.
By training progressively, the method continuously refreshes and reinforces learned representation.

\subsubsection{\textbf{Inference}}
\label{sec:inference}
During inference, we disable the QCCL module (Sec.~\ref{QCCL}). 
In Multi-Granularity Proposal Generation (Sec.~\ref{Multi-Granularity Proposal Generation}), the coarse-grained branch is canceled. 
Inference relies solely on fine-grained representation, thereby preserving the knowledge and performance improvements from the PEHT and allowing for faster, more efficient localization.

\section{Experiments}
\label{EXP}
\begin{table*}[!t] 
	\centering
	\caption{Performance comparison with state-of-the-art methods on the ActivityNet Caption~\cite{ActivityNetCaptions}, Charades-STA~\cite{TALL}, and TACoS~\cite{TACOS} datasets.
The best and second-best results are highlighted in \textbf{bold} and \underline{underlined} text, respectively.}
	\label{tab:sota_comparison}
	\renewcommand{\arraystretch}{1.1}  
	\setlength{\tabcolsep}{1.0pt}     
	\setlength{\arrayrulewidth}{0.4pt}
	\scriptsize  
	\resizebox{\linewidth}{!}{
		\begin{tabular}{l!{\vrule}l!{\vrule} *{3}{c}!{\vrule} *{3}{c}!{\vrule}c!{\vrule} *{3}{c}!{\vrule} *{3}{c}!{\vrule}c!{\vrule} *{3}{c}!{\vrule} *{3}{c}!{\vrule}c}
			\hline  
			\multirow{3}{*}{\textbf{Method}} & \multirow{3}{*}{\textbf{Venue}}  
			& \multicolumn{7}{c!{\vrule}}{\textbf{ActivityNet Caption}} & \multicolumn{7}{c!{\vrule}}{\textbf{Charades-STA}} & \multicolumn{7}{c}{\textbf{TACoS}} \\ 
			\cline{3-23}  
			& & \multicolumn{3}{c!{\vrule}}{\textbf{R@1, IoU@}} & \multicolumn{3}{c!{\vrule}}{\textbf{R@5, IoU@}} & \multirow{2}{*}{\textbf{mIoU}} 
			& \multicolumn{3}{c!{\vrule}}{\textbf{R@1, IoU@}} & \multicolumn{3}{c!{\vrule}}{\textbf{R@5, IoU@}} & \multirow{2}{*}{\textbf{mIoU}} 
			& \multicolumn{3}{c!{\vrule}}{\textbf{R@1, IoU@}} & \multicolumn{3}{c!{\vrule}}{\textbf{R@5, IoU@}} & \multirow{2}{*}{\textbf{mIoU}} \\
			& & \textbf{0.3} & \textbf{0.5} & \textbf{0.7} & \textbf{0.3} & \textbf{0.5} & \textbf{0.7}
			& & \textbf{0.3} & \textbf{0.5} & \textbf{0.7} & \textbf{0.3} & \textbf{0.5} & \textbf{0.7}
			& & \textbf{0.3} & \textbf{0.5} & \textbf{0.7} & \textbf{0.3} & \textbf{0.5} & \textbf{0.7} & \\
			\midrule  
			\multicolumn{23}{@{}l}{\textit{non GCN-based methods}} \\ 
			\midrule
			
			2D-TAN~\cite{2D-TAN} & \textit{AAAI\textquotesingle20} & 59.45 & 44.51 & 26.54 & 85.53 & 77.13 & 61.96 & --- & --- & 39.70 & 23.31 & --- & 80.32 & 51.26 & --- & 37.29 & 25.32 & --- & 57.81 & 45.04 & --- & --- \\
			
			CDN~\cite{CDN} & \textit{TMM\textquotesingle22} & --- & --- & --- & --- & --- & --- & --- & --- & 45.24 & 26.99 & --- & 81.18 & 57.47  & --- & 43.09 & 32.82 & --- & 64.98 & 52.96 & --- & --- \\

			LOCFORMER~\cite{METML} & \textit{EACL\textquotesingle23} & \underline{60.61} & 43.74 & 27.04 & --- & --- & --- & 44.05 & 71.88 & 58.52 & 38.51 & --- & --- & --- & 51.76 & --- & --- & --- & --- & --- & --- & --- \\
			
			PLN~\cite{PLN} & \textit{TOMM\textquotesingle23} & 59.65 & \textbf{45.66} & \textbf{29.28} & 85.66 & 76.65 & \underline{63.06} & \underline{44.12} & 68.60 & 56.02 & 35.16 & \underline{94.54} & \underline{87.63} & \underline{62.34} & 49.09 & 43.89 & 31.12 & --- & 65.11 & 52.89 & --- & 29.70 \\

			R\textsuperscript{2}Tuning~\cite{R2Tuning} & \textit{ECCV\textquotesingle24} & --- & --- & --- & --- & --- & --- & --- & 70.91 & 59.78 & 37.02 & --- & --- & --- & 50.86 & 49.71 & 38.72 & 25.12 & --- & --- & --- & 35.92 \\
			

			NumPro~\cite{NumPro} & \textit{CVPR\textquotesingle25}  & 55.60 & 37.50 & 20.60 & --- & --- & --- & 38.80 & 63.80 & 42.00 & 20.60 & --- & --- & --- & 41.40 & --- & --- & --- & --- & --- & --- & --- \\
			
			LMR~\cite{LMR} & \textit{TMM\textquotesingle25}  & --- & --- & --- & --- & --- & --- & --- & --- & 55.91 & 35.19 & --- & 83.79 & 50.48 & --- & \underline{53.24} & 37.16 & 22.81 & --- & --- & --- & 37.25 \\

			MKBP~\cite{MKBP} & \textit{TCSVT\textquotesingle26}  & --- & --- & --- & --- & --- & --- & --- & 71.77 & \underline{60.43} & \underline{40.81} & --- & --- & --- & 52.05 & 52.71 & \underline{41.99} & \underline{27.63} & --- & --- & --- & \underline{38.34} \\
			
			\midrule
			\multicolumn{23}{@{}l}{\textit{GCN-based methods}} \\ 
			\midrule
			
			DORi~\cite{DORi} & \textit{WACV\textquotesingle21} & 57.89 & 41.35 & 26.41 & --- & --- & --- & 42.79 & \underline{72.72} & 59.65 & 40.56 & --- & --- & --- & \textbf{53.28} & 31.80 & 28.69 & 24.91 & --- & --- & --- & 26.42 \\
			
			PDIN~\cite{PDIN} & \textit{IoTJ\textquotesingle25} & 57.71 & 41.64 & 26.82 & --- & --- & --- & --- & 72.42 & 58.69 & \textbf{42.37} & --- & --- & --- & --- & --- & --- & --- & --- & --- & --- & --- \\
			
			UniSDNet*~\cite{UniSDNet} & \textit{TPAMI\textquotesingle25} & 60.39 & 44.69 & 26.45 & \underline{86.34} & \underline{77.17} & 61.49 & 43.80 & 60.99 & 47.58 & 29.52 & 93.06 & 84.52 & \textbf{63.82} & 43.79 & 49.41 & 34.77 & 17.57 & \underline{73.31} & \underline{57.54} & \underline{29.94} & 34.22 \\
			
			SDGAN (Ours) & --- & \textbf{63.10} & \underline{44.78} & \underline{27.61} & \textbf{87.65} & \textbf{79.65} & \textbf{65.69} & \textbf{45.63} & \textbf{74.44} & \textbf{61.56} & 38.20 & \textbf{97.77} & \textbf{91.56} & 60.48 & \underline{53.18} & \textbf{58.89} & \textbf{44.04} & \textbf{27.87} & \textbf{83.78} & \textbf{72.58} & \textbf{45.09} & \textbf{42.73} \\
			\hline  
		\end{tabular}
	}
\end{table*}

\subsection{Experimental Setup}
\subsubsection{\textbf{Datasets}}
We evaluate the proposed Static and Dynamic Graph Alignment Network (SDGAN) on three widely used Temporal Video Grounding (TVG) benchmark datasets: ActivityNet Caption~\cite{ActivityNetCaptions}, Charades with Sentence Temporal Annotations (Charades-STA)~\cite{TALL}, and Textually Annotated Cooking Scenes (TACoS)~\cite{TACOS}.  
All three datasets consist of real-world videos paired with natural language queries. 
ActivityNet Captions is a large-scale dataset with diverse real-world activities, 
Charades-STA focuses on indoor daily activities, while TACoS consists of long cooking videos.

\subsubsection{\textbf{Evaluation Metrics}}
Following prior works~\cite{2D-TAN, UniSDNet}, we adopt two standard evaluation metrics: $\mathrm{R@}h, \mathrm{IoU@}u$ and $\mathrm{mIoU}$.
$\mathrm{R@}h, \mathrm{IoU@}u$ is the percentage of queries for which at least one of the top-$h$ predicted moments has an Intersection over Union (IoU) $\geq u$ with the ground truth.
$\mathrm{mIoU}$ is computed as the average IoU between the top-1 predicted moment and the corresponding ground-truth  over all queries.

\subsubsection{\textbf{Implementation Details}} 
We implemented the proposed SDGAN in PyTorch and conducted all experiments on a single NVIDIA GeForce RTX 4090D GPU. 
For feature extraction, we employed 3D ConvNet (C3D)~\cite{C3D} or Two-Stream Inflated 3D ConvNet (I3D)~\cite{I3D} as the backbone network to capture dynamic features, depending on the characteristics of each dataset. 
Static features were extracted using a pre-trained Contrastive Language--Image Pre-training (CLIP)~\cite{CLIP} model by sampling one frame every 16 frames, yielding 768-dimensional feature vectors. 
Furthermore, textual queries were encoded using pre-trained Global Vectors for word representation (GloVe)~\cite{GloVe} embeddings. 
The model was trained using the AdamW optimizer~\cite{Adam}. 
Detailed hyperparameter settings and the complete training configurations are summarized in Table~\ref{tab:hyperparam_training}. 
All hyperparameters and the full source code are publicly available at: \url{https://github.com/ZhanJieHu/SDGAN}.

\subsection{Comparison with State-of-the-Art Methods}
We compare the proposed SDGAN with state-of-the-art methods on three widely used TVG benchmark datasets, as summarized in Table~\ref{tab:sota_comparison}. Overall, SDGAN achieved the best performance on the vast majority of evaluation metrics. 
Notably, on the TACoS~\cite{TACOS} dataset, SDGAN outperformed the previous state-of-the-art method UniSDNet*~\cite{UniSDNet} by approximately 15\% in terms of Rank5@0.5 and Rank5@0.7. Moreover, on the large-scale ActivityNet Captions~\cite{ActivityNetCaptions} dataset, SDGAN also surpassed the top-ranked Progressive Localization Network (PLN)~\cite{PLN} in terms of mIoU, demonstrating its ability to produce more precise temporal predictions. 

Compared with non-GCN-based methods, SDGAN achieved better results because GCN-based models provide a more effective mechanism for capturing contextual interactions~\cite{MOC-CRGMN}. Furthermore, we compare SDGAN with other GCN-based methods, including Discovering Object Relationships (DORi)~\cite{DORi}, Progressive Dynamic Interaction Network (PDIN)~\cite{PDIN}, and Unified Static and Dynamic Network (UniSDNet)~\cite{UniSDNet}. Although DORi employs a pre-trained object detector and PDIN incorporates additional audio features, SDGAN still achieved better performance. This advantage can be attributed to the fact that SDGAN learns highly query-aware video representations, while Progressive Easy-to-Hard Training strategy (PEHT) guides the model toward a more effective optimization trajectory.
The results of UniSDNet*~\cite{UniSDNet} were reproduced using its publicly available source code at \url{https://github.com/xian-sh/UniSDNet}. 

\subsection{Ablation Studies}
\label{Ablation Studies}
To systematically evaluate the effectiveness of the proposed SDGAN, we conduct a series of ablation studies on effects of static and dynamic features (Sec.~\ref{Effects of Static and Dynamic Features}), effects of graphs construction (Sec.~\ref{Effects of Graphs Construction}), Effects of Position-wise Nodes Alignment (PNA) (Sec.~\ref{Effects of Position-wise Nodes Alignment (PNA)}), and effects of training Strategy (Sec.~\ref{Effects of Training Strategy}) on the TACoS~\cite{TACOS} dataset.

\subsubsection{\textbf{Effects of Static and Dynamic Features}}
\label{Effects of Static and Dynamic Features}
\begin{figure}[!t]
	\centering
	\includegraphics[width=\columnwidth]{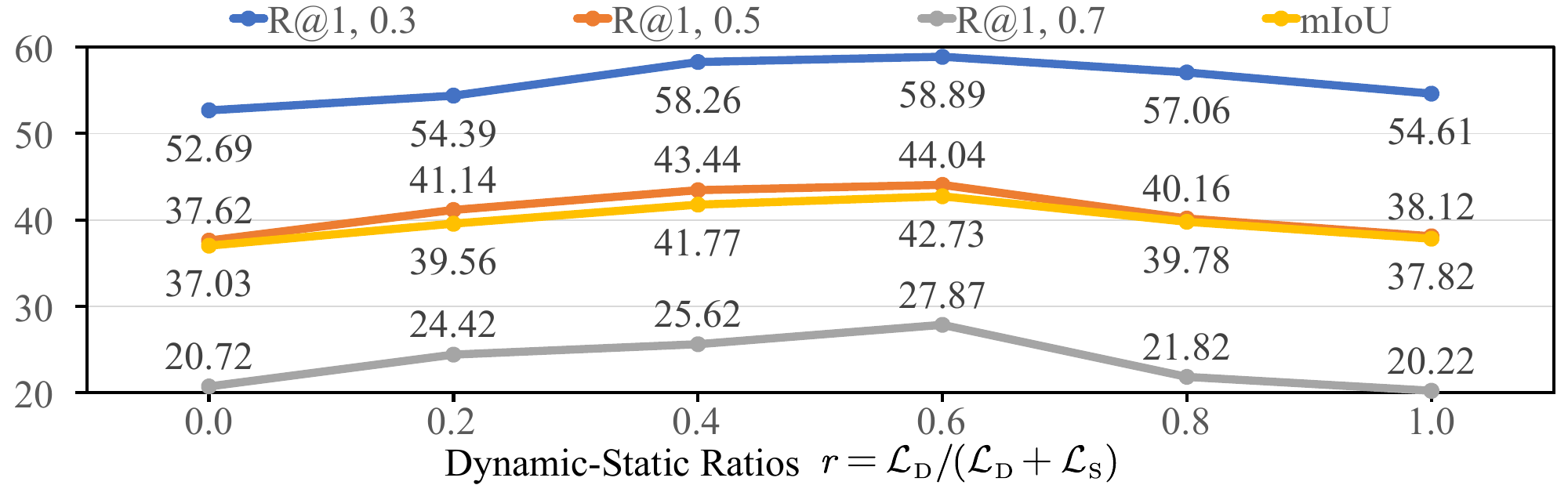}
	\caption{Performance of evaluation metrics under different ratios of dynamic and static features.}
	\label{fig:v&f}
	\vspace{-10pt}
\end{figure}

To investigate the contribution of dynamic and static features, we evaluate the model performance across a range of dynamic--static ratios.
For this, $\mathcal{L}_{\mathrm{D}}$ and $\mathcal{L}_{\mathrm{S}}$ (Eq.~\ref{eq:final_loss}) are employed to regulate the contributions of each feature, respectively. 
The ratio is defined as $r = \mathcal{L}_{\mathrm{D}}/(\mathcal{L}_{\mathrm{D}} + \mathcal{L}_{\mathrm{S}})$, which represents the different proportions of dynamic and static features integrated into the proposed model. 
Specifically, $r = 0$ indicates the exclusive use of static feature, and $r = 1$ indicates the exclusive use of dynamic feature.
The results are reported in Fig.~\ref{fig:v&f}. 
When combining static and dynamic features (when $0<r<1$), all evaluation metrics further improved compared with using static or dynamic features alone (when $r\in\{0,1\}$). 
We can see that the proposed model achieved the best results in all evaluation metrics when the dynamic-static ratio was 0.6.  
These results demonstrate that combining static and dynamic visual features yields more expressive and robust video representations. 
Moreover, dynamic features appeared to contribute slightly more than static features.

\subsubsection{\textbf{Effects of Graphs Construction}}
\label{Effects of Graphs Construction}
As described in Sec.~\ref{QCCL}, Query--Clip Contrastive Learning (QCCL) provides a strong foundation for enabling selective inter-node interaction.
Furthermore, as elaborated in Sec.~\ref{AGM}, Adaptive Graph Modeling (AGM) constructs selective and adaptive edges, departing from the fixed rules adopted in prior work by the Hu et al.~\cite{UniSDNet}.
To evaluate the individual and combined contributions of QCCL and AGM, we conduct ablation studies by removing each component separately.
Specifically, QCCL is disabled or enabled by setting $\lambda_\mathrm{Q}$ (Eq.~\ref{eq:final_loss}) to 0 or 1, respectively. 
The results are summarized in Table~\ref{tab:QCCL_pos}.
Exp.~2 outperforms Exp.~1 by 1.30\% and 0.93\% in R@1, IoU@0.3 and R@1, IoU@0.7, clearly demonstrating the effectiveness of QCCL. 
Exp.~3 achieved improvements of 1.23\% and 0.91\% than Exp.~1 on R@1, IoU@0.5 and mIoU, clearly demonstrating the effectiveness of learnable inter-node interaction in AGM. 
When both strategies are incorporated together, performance of Exp.~4 improved across all metrics. 
These results owe to the fact that QCCL helps generate query-aware video representation, while AGM significantly reduces unnecessary interaction.

\begin{table}[!t]
	\centering
	\caption{ 
Ablation studies on graph construction and Position-wise Nodes Alignment (PNA) on the TACoS~\cite{TACOS} dataset.
		FPNA denotes Fine-grained PNA,
		CPNA denotes Coarse-grained PNA,
		w/ denotes with a specific module,
		Full model denotes the model with all modules enabled.  
		The best results are highlighted in \textbf{Bold} text.}
	\label{tab:QCCL_pos}
	\renewcommand{\arraystretch}{1.1} 
	\setlength{\tabcolsep}{1.2pt} 
	\scriptsize 
	\begin{tabular}{cl|*{2}{c}|*{2}{c}|*{3}{c}!{\vrule}c} 
		\toprule
		\multirow{2}{*}{\textbf{Experiment}} & \multirow{2}{*}{\textbf{Configuration}} & \multirow{2}{*}{\textbf{QCCL}} & \multirow{2}{*}{\textbf{AGM}} & \multirow{2}{*}{\textbf{FPNA}} & \multirow{2}{*}{\textbf{CPNA}} & \multicolumn{3}{c!{\vrule}}{\textbf{R@1, IoU@}} & \multirow{2}{*}{\textbf{mIoU}} \\
		& & & & & & \textbf{0.3} & \textbf{0.5} & \textbf{0.7} & \\
		
		\midrule
		\multicolumn{10}{@{}l}{\textit{Effects of Graphs Construction}} \\ 
		\midrule
		1 & Baseline & --- & --- & --- & \checkmark  & 56.04 & 41.06 & 22.34 & 39.19 \\
		2 & w/ QCCL & \checkmark & --- & --- & \checkmark  & 57.34 & 40.99 & 23.27 & 39.75 \\
		3 & w/ AGM & --- & \checkmark & --- & \checkmark & 56.91 & 42.29 & 23.02 & 40.10 \\
		\rowcolor{gray!10}
		4 & \textbf{Full model} & \textbf{\checkmark} & \textbf{\checkmark} & \textbf{---} & \textbf{\checkmark} & \textbf{58.89} & \textbf{44.04} & \textbf{27.87} & \textbf{42.73} \\
		
		\midrule
		\multicolumn{10}{@{}l}{\textit{Position-wise Nodes Alignment (PNA)}} \\ 
		\midrule
		5 & Baseline  & \checkmark & \checkmark & --- & ---  & 57.54 & 41.34 & 22.54  & 40.17 \\
		6 & w/ FPNA  & \checkmark & \checkmark & \checkmark & ---  & 57.29 & 41.24 & 23.12  & 40.29 \\
		\rowcolor{gray!10}
		7 & \textbf{w/ CPNA} & \checkmark & \checkmark & --- & \checkmark  & \textbf{58.89} & \textbf{44.04} & \textbf{27.87} & \textbf{42.73}  \\
		8 & Full model  & \checkmark & \checkmark & \checkmark & \checkmark & 57.59 & 41.96  & 24.44 & 41.07 \\
		
		\bottomrule
	\end{tabular}
\end{table}

\subsubsection{\textbf{Effects of Position-Wise Nodes Alignment (PNA)}}
\label{Effects of Position-wise Nodes Alignment (PNA)}
As noted in Sec.~\ref{loss functions}, PNA includes coarse-grained ($\mathcal{L}_{\mathrm{PNA, coarse}}$) and fine-grained ($\mathcal{L}_{\mathrm{PNA, fine}}$) loss terms. 
We evaluate their contributions by setting $\lambda_\mathrm{C}$ and $\lambda_\mathrm{F}$ (Eq.~\ref{eq:final_loss}) to 0 or 1.
The results are summarized in Table~\ref{tab:QCCL_pos}.
Exp.~7, which employs coarse-grained PNA, achieved the best performance across all evaluation metrics. By contrast, Exp.~6, which employs fine-grained PNA, showed performance comparable to that of Exp.~5 without PNA. 
This observation can be explained by the inherent correspondence between static and dynamic features at the same spatial locations. Fine-grained alignment tends to impose excessively strict constraints, which limit the flexibility of feature learning, whereas removing alignment altogether can result in information loss. Therefore, adopting only coarse-grained alignment, while excluding fine-grained alignment, appears to be the most effective strategy.

\subsubsection{\textbf{Effects of Training Strategy}}
\label{Effects of Training Strategy}
To analyze the efficiency of PEHT, we evaluate various training strategies over a total of 140 epochs, adopting different loss functions in different epochs.
As illustrated in Table~\ref{tab:training}, we conduct experiments using four distinct training strategies:
1) Fine-grained branch only (serving as the baseline),
2) Coarse-grained branch for the first 70 epochs, followed by the fine-grained branch for the subsequent 70 epochs,
3) Alternating between the coarse-grained and fine-grained branches every 10 epochs, and 
4) Alternating between the coarse-grained and fine-grained branches every 20 epochs.
The results show that strategy 3) consistently outperformed the baseline strategies 1) and 2) across all evaluation metrics. This suggests that progressive alternating training can continuously reinforce the learned representations and facilitate more effective optimization. In addition, alternating the training granularity every 10 epochs proves more effective than using a longer alternation interval. Therefore, strategy 3) is adopted as the default setting of PEHT.

\begin{table}[!t]
	
	\centering
	\caption{Comparison of different training strategies. The best results are highlighted in \textbf{Bold} text.}
	\captionsetup[subfloat]{labelformat=empty, labelsep=none}
	
	\subfloat[]{ 
	\renewcommand{\arraystretch}{1.1} 
	\setlength{\tabcolsep}{0.9pt} 
	\footnotesize 
	\hspace*{-10pt}  
	\begin{tabular}{c p{152pt} |*{3}{c}!{\vrule}c}
		\toprule
		\multirow{2}{*}{\textbf{}} & \multirow{2}{*}{\textbf{Strategy}} & \multicolumn{3}{c!{\vrule}}{\textbf{R@1, IoU@}} & \multirow{2}{*}{\textbf{mIoU}} \\
		& & \textbf{0.3} & \textbf{0.5} & \textbf{0.7} & \\
		\midrule
		
		1 & Fine-grained branch only (Baseline) 		& 58.16 & 43.74 & 26.49 & 41.55 \\
		2 & Coarse-grained first and fine-grained later & 58.06 & 42.29 & 22.19 & 40.70 \\
		\rowcolor{gray!10}
		\textbf{3} & \textbf{Alternate coarse/fine branch every 10 epochs} & \textbf{58.89} & \textbf{44.04} & \textbf{27.87} & \textbf{42.73} \\
		4 & Alternate coarse/fine branch every 20 epochs & 56.94 & 41.49 & 22.97 & 40.41 \\
		\bottomrule
	\end{tabular}
	}
	
	\vspace{-20pt}
	
	\subfloat[]{ 
		\hspace*{-6pt}  
		\includegraphics[scale=1.01, width=1.01\columnwidth, keepaspectratio]{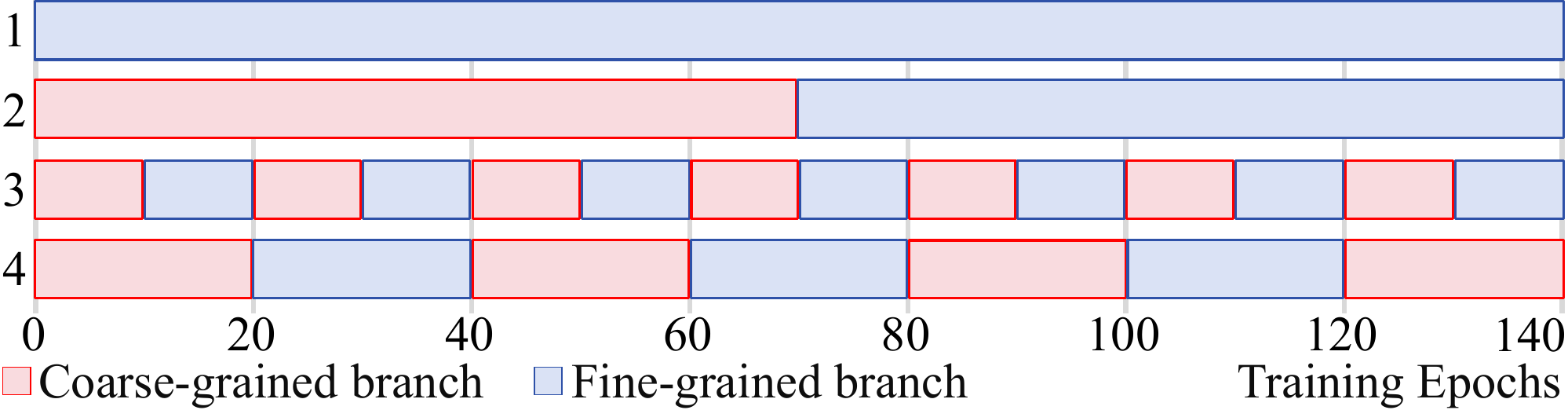}
		\label{fig:training}
	}
	\vspace{-20pt}

	\label{tab:training}
\end{table}

\vspace{-7pt}
\subsection{Qualitative Results}
\begin{figure}[!t]
	\centering
	\captionsetup[subfloat]{labelformat=empty, labelsep=none}
	
	\subfloat[]{ 
		\includegraphics[width=\columnwidth]{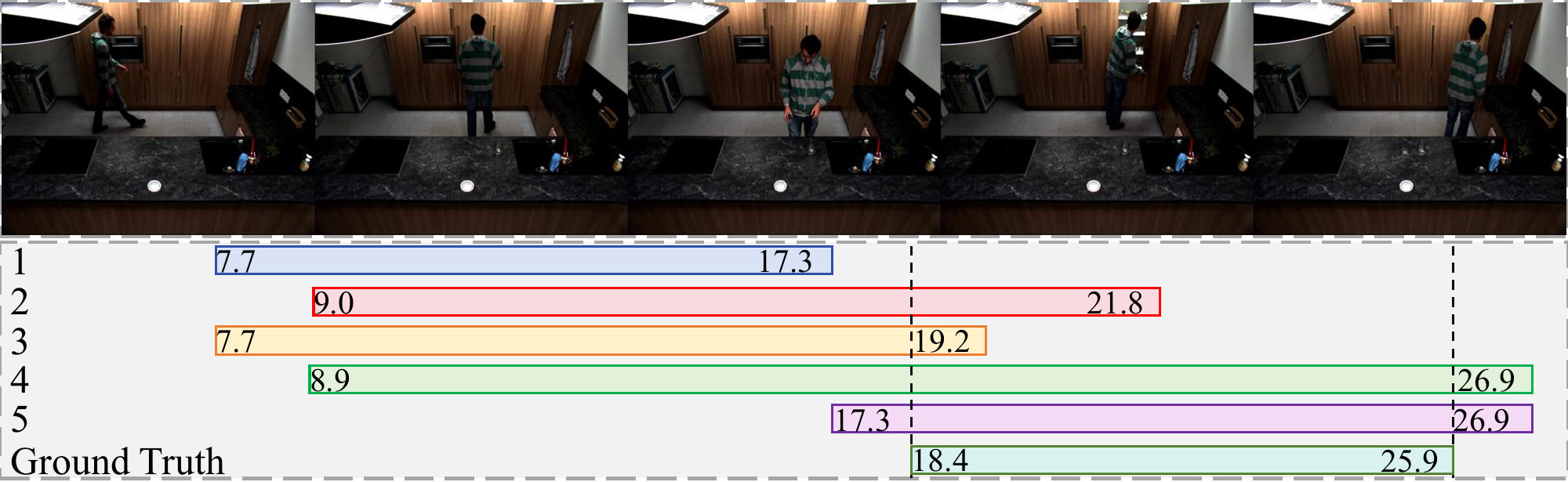}
	}

	\vspace{-22pt}

	\subfloat[]{ 
	\footnotesize 
	\renewcommand{\arraystretch}{1.0} 
	\setlength{\tabcolsep}{1.5pt}
	\hspace*{-8.5pt}  
	\begin{tabularx}{\linewidth}{c|c|c|c| p{167pt}}
		\toprule
		\textbf{Exp.} & \textbf{Feat.} & \textbf{Gph.} & \textbf{Prog.} & \textbf{Explanation} \\
		\midrule
		1 & --- & --- & ---  & Original UniSDNet~\cite{UniSDNet} model \\
		2 & \checkmark & --- & ---  & Both static and dynamic features \\
		3 & --- & \checkmark & --- & The proposed graph building strategies in \ref{DSGN}  \\
		4 & --- & --- & \checkmark & Progressive Easy-to-Hard Training Strategy\\
		\rowcolor{gray!10}
		\textbf{5} & \checkmark & \checkmark & \checkmark & \textbf{Complete SDGAN model}\\
		\bottomrule
	\end{tabularx}
	}
	\vspace{-10pt}
	\caption{Qualitative comparison of different variants on the TACoS~\cite{TACOS} dataset. 
		}
	\label{fig:QualitativeResults}
	\vspace{-10pt}
\end{figure}

Fig.~\ref{fig:QualitativeResults} shows qualitative results on the TACoS~\cite{TACOS} dataset. 
This example is particularly discriminative due to the presence of multiple related queries in a single extended video.

Five model variants are compared.
The plain baseline (Exp.~1) suffers from noticeable semantic bias.
Exp.~2 performed well thanks to the exploitation of complementary static and dynamic features.
Exp.~3 gained improvement through efficient, query-conditioned feature interaction.
Exp.~4 showed strong performance thanks to PEHT that continuously refines the learned representation.

Ultimately, Exp.~5, which combines all proposed components, achieved the most precise target moment localization.
These results strongly confirm the effectiveness of the full SDGAN.

\vspace{-1pt}
\section{Conclusion}
\vspace{-1pt}

In this paper, we proposed Static and Dynamic Graph Alignment Network (SDGAN), a novel framework for Temporal Video Grounding (TVG). SDGAN jointly exploited static and dynamic visual features through Dual-Stream Graph Network (DSGN) and further aligned them via Position-wise Nodes Alignment (PNA) to obtain more expressive video representations. Meanwhile, Query--Clip Contrastive Learning (QCCL) and Adaptive Graph Modeling (AGM) explicitly incorporated query semantics into temporal graph learning, enabling query-aware and targeted feature interaction. In addition, Progressive Easy-to-Hard Training strategy (PEHT) progressively bridged coarse-grained semantic localization and fine-grained temporal boundary refinement, leading to more effective optimization. Extensive experiments on three benchmark datasets demonstrated the superiority of SDGAN and verified the effectiveness of each proposed component. 
In future work, we will further extend SDGAN to more challenging scenarios, such as longer videos and more complex query compositions, while improving its efficiency and generalization.

\section*{Supplementary Material}
The supplementary material is available as a separate PDF on arXiv, including implementation details and an explanation of the notation. It can be accessed via the “ancillary files” section on the arXiv page.

\vspace{-4pt}
\bibliographystyle{IEEEtran}
\bibliography{IEEEabrv,bib}

\end{document}